\documentclass[11pt,a4paper]{article}
\usepackage[hyperref]{acl2019}
\usepackage{times}
\usepackage{latexsym}
\usepackage{graphicx}
\usepackage{tikz-dependency}
\usepackage{enumitem}

\usepackage{url}

\aclfinalcopy 
\def\aclpaperid{BlackboxNLP 30} 

\def\RR#1{{\color{blue}RR: \it #1}}

\def\RR#1{}

\title{Inducing syntactic trees from BERT representations}

\author{Rudolf Rosa \and David Mare\v{c}ek\\
  Charles University, Faculty of Mathematics and Physics \\
  Institute of Formal and Applied Linguistics \\
  Malostransk\' e n\' am\v est\' i 25, 118 00 Prague, Czech Republic \\
  \texttt{\{rosa, marecek\}@ufal.mff.cuni.cz}}
\date{}

\begin{document}
\maketitle

\section{Introduction}

One of the traditional linguistic criteria for recognizing dependency relations
in a sentence is that a head of a syntactic construction determines its syntactic category and can often replace it without any damage of syntactic correctness~\cite{kubler:2009,lopatkova:2005}.
\citet{marecek:2012:emnlp} use similar principle for unsupervised dependency parsing. They estimate ``reducibility'', a property describing how easily a word may be omitted from sentence without damaging it. Their hypothesis is that reducible words are more likely to occur as leaves in dependency trees. 
They simply declare a word reducible if the same sentence without this word appears elsewhere in the corpus. This very sparse method showed that reducibility of adjectives and adverbs is high, whereas reducibility of verbs is quite low.

With the advance of neural-network language models, e.g.\ BERT~\cite{devlin:2018} or ELMo~\cite{peters:2018},
there are new ways how to estimate reducibilities of words. In this paper, we use English model of BERT and explore how a deletion of one word in a sentence changes representations of other words.
Our hypothesis is that removing a reducible word (e.g.\ an adjective) does not affect representation of other words so much as removing e.g.\ the main verb, which makes the sentence ungrammatical and of ``high surprise'' for the language model.

We estimate reducibilities of individual words and also of longer phrases (word n-grams), study their syntax-related properties, and then also use them to induce full dependency trees.
A significant difference between our work and most previous works~\cite{hewitt:2019,belinkov:phd:2018} is that we estimate the reducibilities and dependency trees directly from the models, without any training on syntactically annotated data.


\section{Reducibility scores from BERT}

\RR{TODO obrázek by byl supr}

We use the pretrained English model (BERT-Large, uncased)\footnote{\url{https://storage.googleapis.com/BERT_models/2018_10_18/uncased_L-24_H-1024_A-16.zip}}, and sentences from the development part of the EWT English treebank from Universal Dependencies 2.3 \cite{ud}; we subselect sentences containing only words included in the vocabulary of the BERT model.

We compute the \textit{reducibility} $r_{p,s}$ of each \textit{phrase} $p$ (any word or continuous sequence of words) in each sentence $s$ as the average change of BERT representations of the words in the sentence when the phrase $p$ is removed. By a BERT representation of a word we mean the state on the last layer of the BERT encoder on the position corresponding to the word.
\begin{equation}
    r_{p,s} = \frac{1}{|s_{-p}|} \sum_{w \in s_{-p}} ||b_{w, s} - b_{w, s_{-p}}||_2
\end{equation}
where $s_{-p}$ is the sentence $s$ with phrase $p$ removed, $b_{w, s}$ is the BERT representation of word $w$ in sentence $s$, and the distance of the BERT representations is the Euclidean distance.
The $b_{w, s_{-p}}$ representations are obtained by simply running BERT on the sentence $s$ with phrase $p$ deleted. 

\RR{poznamenat že tohle je spíš sémantická reducibilita, zatimco my "chceme" spíš syntaktickou reducibiklitu?}



\section{Linguistic properties of reducibilities}

\RR{vyhodit nějaký nebo všechny obrázky?}

\begin{figure}
    \centering
    \includegraphics[width=0.5\textwidth]{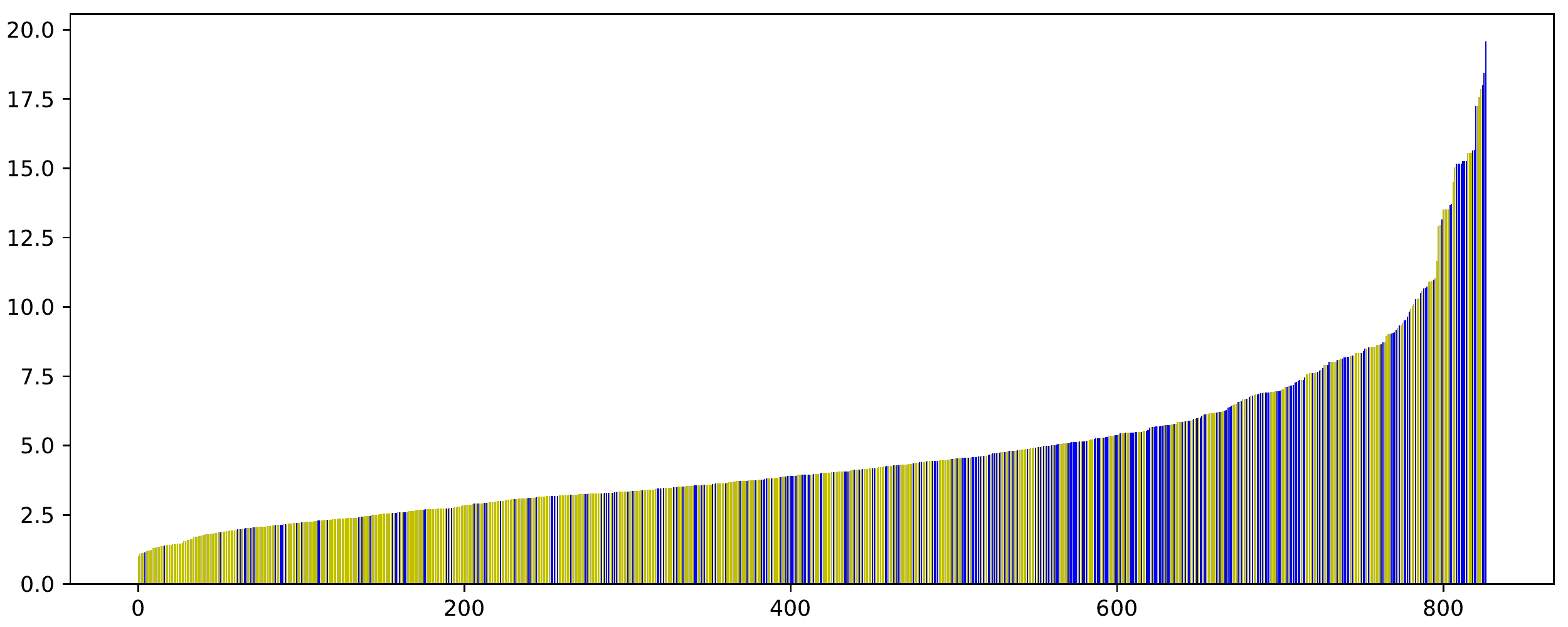}
    \includegraphics[width=0.5\textwidth]{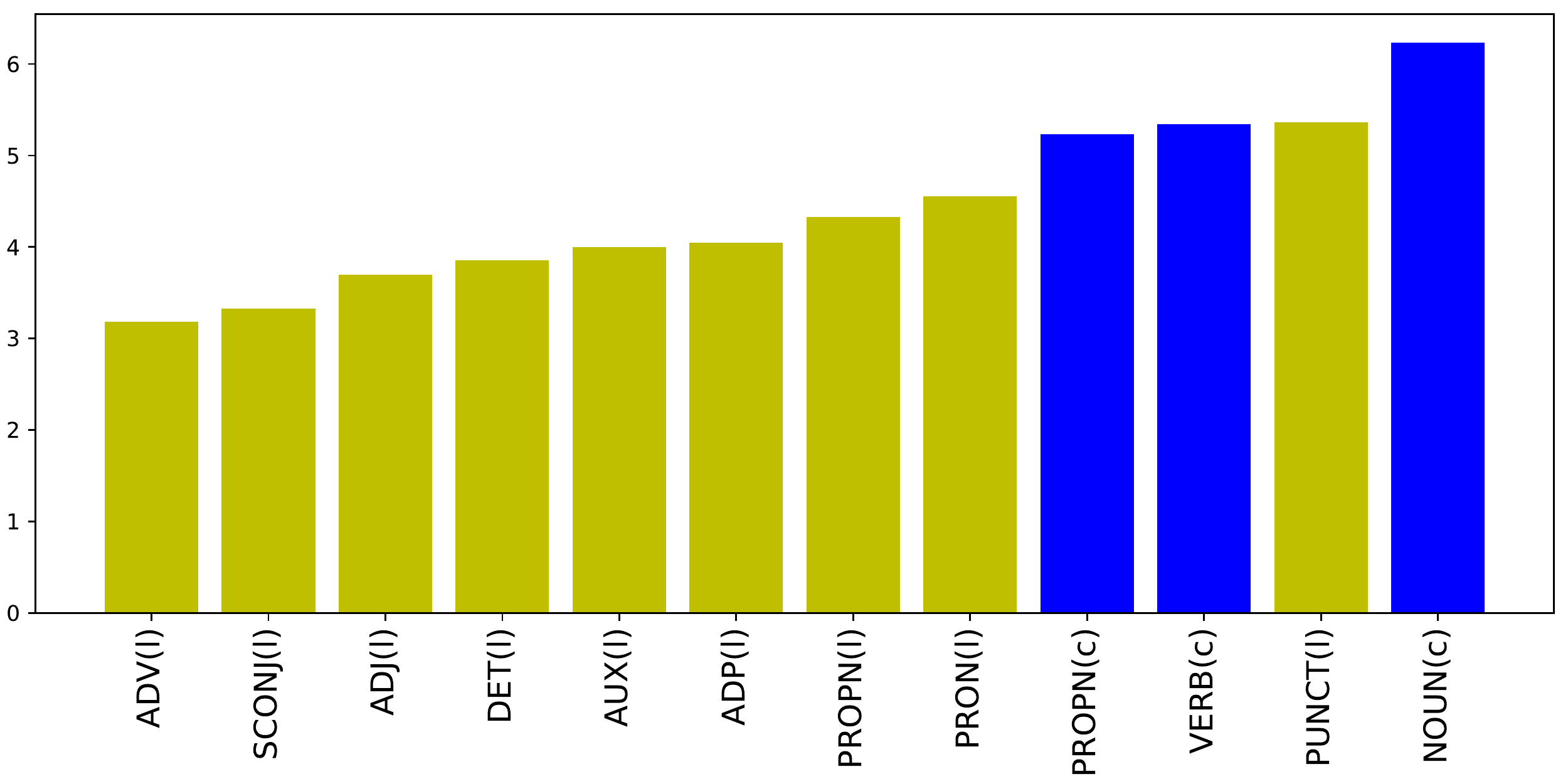}
    \caption{Distribution of reducibility scores across all the tested instances and averaged scores for POS tags. The leaf words are yellow, non-leaf words are blue.}
    \label{fig:reducibilities}
\end{figure}

In Figure~\ref{fig:reducibilities}, we show reducibility scores of all the words in our testing data
and average reducibility scores for individual part-of-speech tags.
To visualise how the word reducibility correlates with being a leaf in the dependency tree, we color all the leaf instances by yellow and all the non-leaf by blue. It is apparent that in the right side of the graph, the blue instances prevail.

The absolute word reducibilities are different in each sentence, but we have found that the threshold separating leafs and non-leaves in a given sentence is around $1.2\times$ the average word reducibility in that sentence. This allows us to separate leafs from non-leaves with an accuracy of 74.5\%, compared to the baseline of 66.4\% (assuming everything is a leaf).


The syntactic root of the sentence tends to be the least reducible word. It is so in 34\% of the sentences; or in 46\% if we ignore punctuation, which tends to be very irreducible.
The random baseline here is 13\%.



The dependency edge direction can be identified with a 70.6\% accuracy, assuming that the parent
should be less reducible than the child node.
The right-chain baseline, assuming that parent is to the right of the child, has a 65.8\% accuracy.

\section{Building dependency trees}

We now examine to which extent the reducibilities extracted from BERT can be used to build dependency trees. We propose two algorithms, and compare them to an uninformed baseline, which
is the right chain, attaching each node to its right neighbor;
the rightmost node becomes the root.
For English, this is quite a strong baseline.


\textbf{Algorithm D:}
In Algorithm D, we construct a projective dependency tree based on phrase reducibilities.
We use the recursive headed brackets encoding, where each subtree is enclosed in one pair of brackets, containing subtrees (phrases in brackets) and just one head (word without brackets), e.g.:
\textit{( (subtree) head (subtree) (subtree) )}.
In each step, we greedily insert a new pair of brackets corresponding to the most reducible phrase such that
the resulting structure still satisfies the following conditions:
(a) brackets do not cross each other
(b) each subtree has a head.
Table~\ref{tab:results} shows that the resulting structures only slightly surpass the uninformed baseline (this can be improved by explicitly setting a low reducibility for punctuation).




\textbf{Algorithm R:}
Algorithm R directly builds upon the right-chain baseline, modifying it by introducing a constraint that the parent of each node must be less reducible than the child node; the least reducible node becomes the root.
Each node is thus attached to the nearest subsequent more reducible node; or to the root if all subsequent nodes are less reducible.
Table~\ref{tab:results} shows that this outperforms the baseline by 7 (or 11) percentage points.





\begin{table}
    \centering
    \begin{tabular}{rr}
         parser & UAS \\
         \hline
         left chain baseline      &  6.8 \\
         right chain baseline     & 29.5 \\
         \hline
         algorithm D              & 31.1 \\
         algorithm D, red. punct. & 33.1 \\
         \hline
         algorithm R              & 37.0 \\
         algorithm R, red. punct. & 40.6 \\
    \end{tabular}
    \caption{Parsing results, Unlabelled Attachment Score}
    \label{tab:results}
\end{table}

\section{Conclusions}

We examine to what extent reducibility, which underlies dependency syntactic structures, can be estimated from BERT representations.
We devise a method based on measuring the differences in the representations when a word or phrase is removed from the sentence, and denoting this difference as the reducibility score of that word or phrase.
We find that such scores partially correspond to the notion of reducibility in dependency trees, seeing a tendency of child nodes and leaf nodes to be more reducible than parent nodes and the root.
We also show that these scores can be used in a simple parsing algorithm to construct dependency trees which are more accurate than an uninformed baseline.


\section*{Acknowledgements}
This work has been supported by the grant 18-02196S of the Czech
Science Foundation and uses language resources and tools developed and
stored by the LINDAT/\discretionary{}{}{}CLARIN project
(LM2015071).

\bibliography{ms}
\bibliographystyle{acl_natbib}

\end{document}